\documentclass[10pt,twocolumn,letterpaper]{article}

\usepackage{cvpr}
\usepackage{times}
\usepackage{epsfig}
\usepackage{graphicx}
\usepackage{amsmath}
\usepackage{amssymb}
\usepackage{nicefrac}
\usepackage{multirow}
\usepackage{array}
\usepackage{color}
\usepackage{bm}
\usepackage{fixltx2e}
\usepackage{algorithm, algpseudocode}

\usepackage{etoolbox}
\makeatletter
\patchcmd\@combinedblfloats{\box\@outputbox}{\unvbox\@outputbox}{}{%
   \errmessage{\noexpand\@combinedblfloats could not be patched}%
}%
 \makeatother

\usepackage[pagebackref=true,breaklinks=true,letterpaper=true,colorlinks,bookmarks=false]{hyperref}

\DeclareMathOperator*{\argmax}{arg\,max} 
\DeclareMathOperator*{\argmin}{arg\,min} 

\cvprfinalcopy 

\def\cvprPaperID{5297} 

\begin{document}

\title{Evading Defenses to Transferable Adversarial Examples by Translation-Invariant Attacks}

\author{Yinpeng Dong, Tianyu Pang, Hang Su, Jun Zhu\thanks{Corresponding author.}\\
Dept. of Comp. Sci. and Tech., BNRist Center, State Key Lab for Intell. Tech. \& Sys.,\\
Institute for AI, THBI Lab, Tsinghua University, Beijing, 100084, China\\
\tt\small{\{dyp17,  pty17\}@mails.tsinghua.edu.cn, \{suhangss, dcszj\}@mail.tsinghua.edu.cn}}

\maketitle

\begin{abstract}
   Deep neural networks are vulnerable to adversarial examples, which can mislead classifiers by adding imperceptible perturbations. An intriguing property of adversarial examples is their good transferability, making black-box attacks feasible in real-world applications. Due to the threat of adversarial attacks, many methods have been proposed to improve the robustness. Several state-of-the-art defenses are shown to be robust against transferable adversarial examples. In this paper, we propose a translation-invariant attack method to generate more transferable adversarial examples against the defense models. By optimizing a perturbation over an ensemble of translated images, the generated adversarial example is less sensitive to the white-box model being attacked and has better transferability. To improve the efficiency of attacks, we further show that our method can be implemented by convolving the gradient at the untranslated image with a pre-defined kernel. Our method is generally applicable to any gradient-based attack method. Extensive experiments on the ImageNet dataset validate the effectiveness of the proposed method. Our best attack fools eight state-of-the-art defenses at an $82\%$ success rate on average based only on the transferability, demonstrating the insecurity of the current defense techniques.
\end{abstract}

\section{Introduction}
\label{sec:intro}

Despite the great success, deep neural networks have been shown to be highly vulnerable to adversarial examples~\cite{Biggio2013Evasion,szegedy2013intriguing,goodfellow2014explaining}.
These maliciously generated adversarial examples are indistinguishable from legitimate ones by adding small perturbations, but make deep models produce unreasonable predictions. 
The existence of adversarial examples, even in the physical world~\cite{kurakin2016adversarial,Eykholt_2018_CVPR,Athalye2017Synthesizing}, has raised concerns in security-sensitive applications, \eg, self-driving cars, healthcare and finance.

\begin{figure}[t]
\centering
\includegraphics[width=0.86\columnwidth]{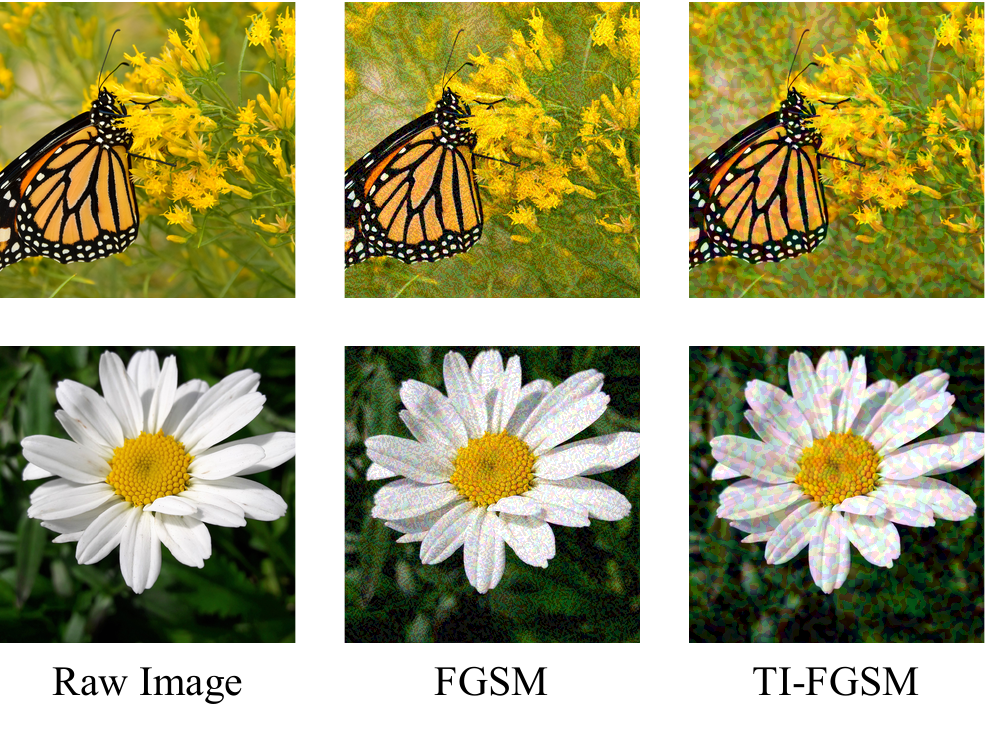}
\caption{The adversarial examples generated by the fast gradient sign method (FGSM)~\cite{goodfellow2014explaining} and the proposed translation-invariant FGSM (TI-FGSM) for the Inception v3~\cite{szegedy2015rethinking} model.}
\label{fig:adv-images}
\vspace{-2ex}
\end{figure}

\begin{figure*}[t]
\centering
\includegraphics[width=0.9\linewidth]{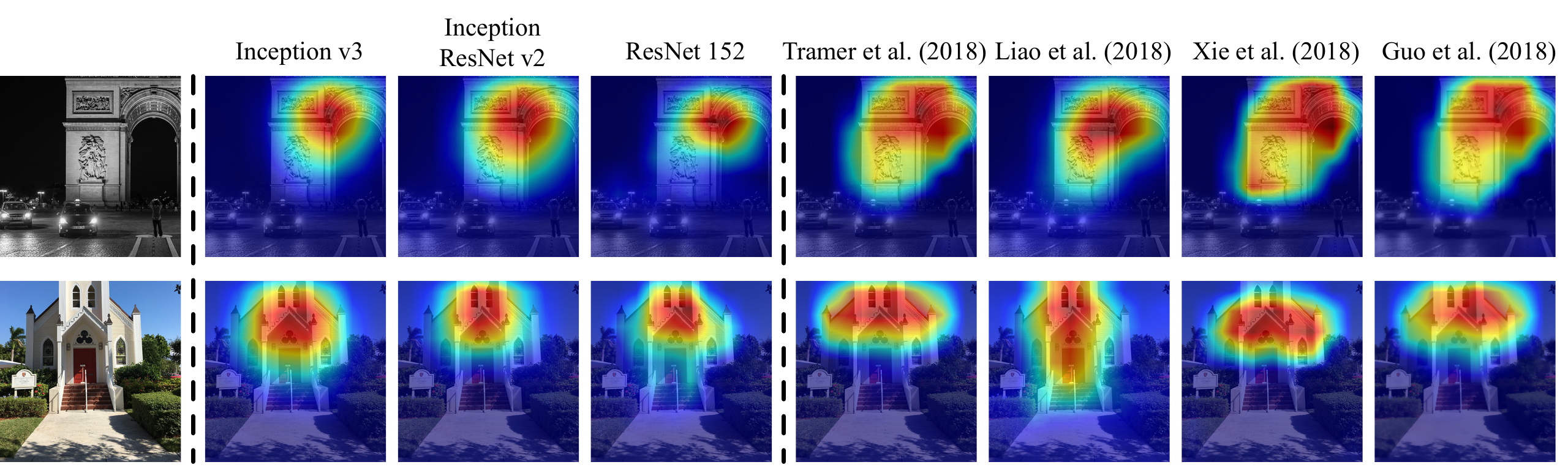}
\caption{Demonstration of the different discriminative regions of the defense models compared with normally trained models. We adopt \textit{class activation mapping}~\cite{zhou2016learning} to visualize the attention maps of three normally trained models---Inception v3~\cite{szegedy2015rethinking}, Inception ResNet v2~\cite{szegedy2017inception}, ResNet 152~\cite{he2015deep} and four defense models~\cite{tramer2017ensemble,Liao2017Defense,Xie2018Mitigating,Guo2017Countering}. These defense models rely on different discriminative regions for predictions compared with normally trained models, which could affect the transferability of adversarial examples.}
\label{fig:attention}
\vspace{-2ex}
\end{figure*}

Attacking deep neural networks has drawn an increasing attention since the generated adversarial examples can serve as an important surrogate to evaluate the robustness of different models~\cite{carlini2017towards} and improve the robustness~\cite{goodfellow2014explaining,Madry2017Towards}.
Several methods have been proposed to generate adversarial examples with the knowledge of the gradient information of a given model, such as fast gradient sign method~\cite{goodfellow2014explaining}, basic iterative method~\cite{kurakin2016adversarial}, and Carlini \& Wagner's method~\cite{carlini2017towards}, which are known as \textit{white-box} attacks.
Moreover, it is shown that adversarial examples have cross-model transferability \cite{liu2016delving}, \ie, the adversarial examples crafted for one model can fool a different model with a high probability.
The transferability enables practical \textit{black-box} attacks to real-world applications and induces serious security issues.

The threat of adversarial examples has motivated extensive research on building robust models or techniques to defend against adversarial attacks. 
These include training with adversarial examples~\cite{goodfellow2014explaining,tramer2017ensemble,Madry2017Towards}, image denoising/transformation~\cite{Liao2017Defense,Xie2018Mitigating,Guo2017Countering}, theoretically-certified defenses~\cite{Raghunathan2018Certified,Wong2018Provable}, and others~\cite{pang2018max,Song2018PixelDefend,Samangouei2018Defense}.
Although the non-certified defenses have demonstrated robustness against common attacks, they do so by causing obfuscated gradients, which can be easily circumvented by new attacks~\cite{Athalye2018Obfuscated}.
However, some of the defenses~\cite{tramer2017ensemble,Liao2017Defense,Xie2018Mitigating,Guo2017Countering} claim to be resistant to transferable adversarial examples, making it difficult to evade them by black-box attacks.

The resistance of the defense models against transferable adversarial examples is largely due to the phenomenon that the defenses make predictions based on different discriminative regions compared with normally trained models.
For example, we show the attention maps of several normally trained models and defense models in Fig.~\ref{fig:attention}, to represent the discriminative regions for their predictions. It can be seen that the normally trained models have similar attention maps while the defenses induce different attention maps.
A similar observation is also found in~\cite{tsipras2018robustness} that the gradients of the defenses in the input space align well with human perception, while those of normally trained models appear very noisy.
This phenomenon of the defenses is caused by either training under different data distributions~\cite{tramer2017ensemble} or transforming the inputs before classification~\cite{Liao2017Defense,Xie2018Mitigating,Guo2017Countering}.
For black-box attacks based on the transferability~\cite{goodfellow2014explaining,liu2016delving,Dong_2018_CVPR}, an adversarial example is usually generated for a single input against a white-box model. So the generated adversarial example is highly correlated with the discriminative region or gradient of the white-box model at the given input point, making it hard to transfer to other defense models that depend on different regions for predictions.
Therefore, the transferability of adversarial examples is largely reduced to the defenses.

To mitigate the effect of different discriminative regions between models and evade the defenses by transferable adversarial examples, we propose a \textbf{translation-invariant attack} method. 
In particular, we generate an adversarial example for an ensemble of images composed of a legitimate one and its translated versions.
We expect that the resultant adversarial example is less sensitive to the discriminative region of the white-box model being attacked, and has a higher probability to fool another black-box model with a defense mechanism.
However, to generate such a perturbation, we need to calculate the gradients for all images in the ensemble, which brings much more computations.
To improve the efficiency of our attacks, we further show that our method can be implemented by convolving the gradient at the untranslated image with a pre-defined kernel under a mild assumption.
By combining the proposed method with any gradient-based attack method (\eg, fast gradient sign method~\cite{goodfellow2014explaining}, \etc), we obtain more transferable adversarial examples with similar computation complexity.

Extensive experiments on the ImageNet dataset~\cite{russakovsky2015imagenet} demonstrate that the proposed translation-invariant attack method helps to improve the success rates of black-box attacks against the defense models by a large margin.
Our best attack reaches an average success rate of $82\%$ to evade eight state-of-the-art defenses based only on the transferability, thus demonstrating the insecurity of the current defenses.

\section{Related Work}
\label{sec:related-work}
\textbf{Adversarial examples.} 
Deep neural networks have been shown to be vulnerable to adversarial examples first in the visual domain~\cite{szegedy2013intriguing}. 
Then several methods are proposed to generate adversarial examples for the purpose of high success rates and minimal size of perturbations~\cite{goodfellow2014explaining,kurakin2016adversarial,carlini2017towards}.
They also exist in the physical world~\cite{kurakin2016adversarial,Eykholt_2018_CVPR,Athalye2017Synthesizing}.
Although adversarial examples are recently crafted for many other domains, we focus on image classification tasks in this paper.

\textbf{Black-box attacks.}
Black-box adversaries have no access to the model parameters or gradients.
The transferability~\cite{liu2016delving} of adversarial examples can be used to attack a black-box model.
Several methods~\cite{Dong_2018_CVPR,Xie2018Improving} have been proposed to improve the transferability, which enable powerful black-box attacks.
Besides the transfer-based black-box attacks, there is another line of work that performs attacks based on adaptive queries.
For example, Papernot \etal~\cite{papernot2016practical} use queries to distill the knowledge of the target model and train a surrogate model. They therefore turn the black-box attacks to the white-box attacks. 
Recent methods use queries to estimate the gradient or the decision boundary of the black-box model~\cite{Chen2017ZOO,Brendel2018Decision} to generate adversarial examples.
However, these methods usually require a large number of queries, which is impractical in real-world applications.
In this paper, we resort to transfer-based black-box attacks.

\textbf{Attacks for an ensemble of examples.} An adversarial perturbation can be generated for an ensemble of legitimate examples. In~\cite{Moosavi2016Universal}, the universal perturbations are generated for the entire data distribution, which can fool the models on most of natural images.
In~\cite{Athalye2017Synthesizing}, the adversarial perturbation is optimized over a distribution of transformations, which is similar to our method. The major difference between the method in~\cite{Athalye2017Synthesizing} and ours is three-fold. First, we want to generate transferable adversarial examples against the defense models, while the authors in~\cite{Athalye2017Synthesizing} propose to synthesize robust adversarial examples in the physical world. Second, we only use the translation operation, while they use a lot of transformations such as rotation, translation, addition of noise, \etc. Third, we develop an efficient algorithm for optimization that only needs to calculate the gradient for the untranslated image, while they calculate the gradients for a batch of transformed images by sampling.

\textbf{Defend against adversarial attacks.}
A large variety of methods have been proposed to increase the robustness of deep learning models.
Besides directly making the models produce correct predictions for adversarial examples, some methods attempt to detect them instead~\cite{Metzen2017On,Pang2018Towards}.
However, most of the non-certified defenses demonstrate the robustness by causing obfuscated gradients, which can be successfully circumvented by new attacks~\cite{Athalye2018Obfuscated}.
Although these defenses are not robust in the white-box setting, some of them~\cite{tramer2017ensemble,Liao2017Defense,Xie2018Mitigating,Guo2017Countering} empirically show the resistance against transferable adversarial examples in the black-box setting.
In this paper, we focus on generating more transferable adversarial examples against these defenses.

\section{Methodology}
\label{sec:method}

In this section, we provide the detailed description of our algorithm.
Let $\bm{x}^{real}$ denote a real example and $y$ denote the corresponding ground-truth label. Given a classifier $f(\bm{x}):\mathcal{X}\rightarrow\mathcal{Y}$ that outputs a label as the prediction for an input, we want to generate an adversarial example $\bm{x}^{adv}$ which is visually indistinguishable from $\bm{x}^{real}$ but fools the classifier, \ie, $f(\bm{x}^{adv})\neq y$.\footnote{This corresponds to untargeted attack. The method in this paper can be simply extended to targeted attack.}
In most cases, the $L_p$ norm of the adversarial perturbation is required to be smaller than a threshold $\epsilon$ as $||\bm{x}^{adv} - \bm{x}^{real}||_p \leq \epsilon$.
In this paper, we use the $L_{\infty}$ norm as the measurement.
For adversarial example generation, the objective is to maximize the loss function $J(\bm{x}^{adv}, y)$ of the classifier, where $J$ is often the cross-entropy loss. So the constrained optimization problem can be written as
\begin{equation}
    \argmax_{\bm{x}^{adv}} J(\bm{x}^{adv}, y), \quad \text{s.t.} \; \|\bm{x}^{adv} - \bm{x}^{real}\|_{\infty} \leq \epsilon.
    \label{eq:problem}
\end{equation}
To solve this optimization problem, the gradient of the loss function with respect to the input needs to be calculated, termed as white-box attacks.
However, in some cases, we cannot get access to the gradients of the classifier, where we need to perform attacks in the black-box manner. We resort to transferable adversarial examples which are generated for a different white-box classifier but have high transferability for black-box attacks.

\subsection{Gradient-based Adversarial Attack Methods}
\label{sec:3.1}
Several methods have been proposed to solve the optimization problem in Eq.~\eqref{eq:problem}. We give a brief introduction of them in this section.

\textbf{Fast Gradient Sign Method (FGSM)}~\cite{goodfellow2014explaining} generates an adversarial example $\bm{x}^{adv}$ by linearizing the loss function in the input space and performing one-step update as
\begin{equation}
    \bm{x}^{adv} = \bm{x}^{real} + \epsilon\cdot\mathrm{sign}(\nabla_{\bm{x}}J(\bm{x}^{real},y)),
\end{equation}
where $\nabla_{\bm{x}}J$ is the gradient of the loss function with respect to $\bm{x}$. $\mathrm{sign}(\cdot)$ is the sign function to make the perturbation meet the $L_{\infty}$ norm bound. FGSM can generate more transferable adversarial examples but is usually not effective enough for attacking white-box models~\cite{Kurakin2017Adversarial}.

\textbf{Basic Iterative Method (BIM)}~\cite{kurakin2016adversarial} extends FGSM by iteratively applying gradient updates multiple times with a small step size $\alpha$, which can be expressed as
\begin{equation}
\bm{x}_{t+1}^{adv} = \bm{x}_t^{adv} + \alpha\cdot\mathrm{sign}(\nabla_{\bm{x}}J(\bm{x}_t^{adv},y)),
\label{eq:iter}
\end{equation}
where $\bm{x}_0^{adv} = \bm{x}^{real}$.
To restrict the generated adversarial examples within the $\epsilon$-ball of $\bm{x}^{real}$, we can clip $\bm{x}_t^{adv}$ after each update, or set $\alpha=\nicefrac{\epsilon}{T}$, with $T$ being the number of iterations.
It has been shown that BIM induces much more powerful white-box attacks than FGSM at the cost of worse transferability~\cite{Kurakin2017Adversarial,Dong_2018_CVPR}.

\textbf{Momentum Iterative Fast Gradient Sign Method (MI-FGSM)}~\cite{Dong_2018_CVPR} proposes to improve the transferability of adversarial examples by integrating a momentum term into the iterative attack method. The update procedure is
\begin{equation}
    \bm{g}_{t+1} = \mu \cdot \bm{g}_{t} + \frac{\nabla_{\bm{x}}J(\bm{x}_{t}^{adv},y)}{\|\nabla_{\bm{x}}J(\bm{x}_{t}^{adv},y)\|_1},
\end{equation}
\begin{equation}
   \bm{x}_{t+1}^{adv} = \bm{x}_{t}^{adv} + \alpha\cdot\mathrm{sign}(\bm{g}_{t+1}),
\end{equation}
where $\bm{g}_t$ gathers the gradient information up to the $t$-th iteration with a decay factor $\mu$.

\textbf{Diverse Inputs Method}~\cite{Xie2018Improving} applies random transformations to the inputs and feeds the transformed images into the classifier for gradient calculation. The transformation includes random resizing and padding with a given probability. This method can be combined with the momentum-based method to further improve the transferability.

\textbf{Carlini \& Wagner's method (C\&W)}~\cite{carlini2017towards} is a powerful optimization-based method, which solves 
\begin{equation}
\argmin_{\bm{x}^{adv}}\| \bm{x}^{adv}- \bm{x}^{real}\|_p 
- c \cdot J(\bm{x}^{adv}, y),
\end{equation}
where the loss function $J$ could be different from the cross-entropy loss. This method aims to find adversarial examples with minimal size of perturbations, to measure the robustness of different models. It also lacks the effectiveness for black-box attacks like BIM.

\subsection{Translation-Invariant Attack Method}

Although many attack methods~\cite{Dong_2018_CVPR,Xie2018Improving} can generate adversarial examples with very high transferability across normally trained models, they are less effective to attack defense models in the black-box manner. Some of the defenses~\cite{tramer2017ensemble,Liao2017Defense,Xie2018Mitigating,Guo2017Countering} are shown to be quite robust against black-box attacks. So we want to answer that: \textit{Are these defenses really free from transferable adversarial examples?}

We find that the discriminative regions used by the defenses to identify object categories are  different from those used by normally trained models, as shown in Fig.~\ref{fig:attention}.
When generating an adversarial example by the methods introduced in Sec.~\ref{sec:3.1}, the adversarial example is only optimized for a single legitimate example.
So it may be highly correlated with the discriminative region or gradient of the white-box model being attacked at the input data point.
For other black-box defense models that have different discriminative regions or gradients, the adversarial example can hardly remain adversarial.
Therefore, the defenses are shown to be robust against transferable adversarial examples.

To generate adversarial examples that are less sensitive to the discriminative regions of the white-box model, we propose a \textbf{translation-invariant attack} method.
In particular, rather than optimizing the objective function at a single point as Eq.~\eqref{eq:problem}, the proposed method uses a set of translated images to optimize an adversarial example as
\begin{equation}
\begin{gathered}
    \argmax_{\bm{x}^{adv}} \sum_{i,j} w_{ij} J(T_{ij}(\bm{x}^{adv}), y), \\ \text{s.t.} \; \|\bm{x}^{adv} - \bm{x}^{real}\|_{\infty} \leq \epsilon,
    \label{eq:problem-new}
    \end{gathered}
\end{equation}
where $T_{ij}(\bm{x})$ is the translation operation that shifts image $\bm{x}$ by $i$ and $j$ pixels along the two-dimensions respectively, \ie, each pixel $(a,b)$ of the translated image is $T_{ij}(\bm{x})_{a,b} = x_{a-i,b-j}$, and $w_{ij}$ is the weight for the loss $J(T_{ij}(\bm{x}^{adv}), y)$. We set $i, j \in \{-k, ..., 0, ..., k\}$ with $k$ being the maximal number of pixels to shift.
With this method, the generated adversarial examples are less sensitive to the discriminative regions of the white-box model being attacked, which may be transferred to another model with a higher success rate.
We choose the translation operation in this paper rather than other transformations (\eg, rotation, scaling, \etc), because we can develop an efficient algorithm to calculate the gradient of the loss function by the assumption of the translation-invariance~\cite{LeCun1995Convolutional} in convolutional neural networks.

\vspace{-1.8ex}
\subsubsection{Gradient Calculation}
\label{sec:gradient}
To solve the optimization problem in Eq.~\eqref{eq:problem-new}, we need to calculate the gradients for $(2k+1)^2$ images, which introduces much more computations.
Sampling a small number of translated images for gradient calculation is a feasible way~\cite{Athalye2017Synthesizing}.
But we show that we can calculate the gradient for only one image under a mild assumption.

Convolutional neural networks are supposed to have the translation-invariant property~\cite{LeCun1995Convolutional}, that an object in the input can be recognized in spite of its position.
In practice, CNNs are not truly translation-invariant~\cite{goodfellow2009measuring,kauderer2017quantifying}.
So we make an assumption that the translation-invariant property is nearly held with very small translations (which is empirically validated in Sec.~\ref{sec:assumption}).
In our problem, we shift the image by no more than 10 pixels along each dimension (\ie, $k \leq 10$).
Therefore, based on this assumption, the translated image $T_{ij}(\bm{x})$ is almost the same as $\bm{x}$ as inputs to the models, as well as their gradients
\begin{equation}
\nabla_{\bm{x}}J(\bm{x}, y)\big|_{\bm{x}=T_{ij}(\hat{\bm{x}})} \approx \nabla_{\bm{x}}J(\bm{x}, y) \big|_{\bm{x}=\hat{\bm{x}}}.
\label{eq:gradient-2}
\end{equation}
We then calculate the gradient of the loss function defined in Eq.~\eqref{eq:problem-new} at a point $\hat{\bm{x}}$ as
\begin{equation}
\begin{split}
        &\nabla_{\bm{x}} \big(\sum_{i,j} w_{ij} J(T_{ij}(\bm{x}), y)\big)\big|_{\bm{x}=\hat{\bm{x}}}\\
        =&\sum_{i,j} w_{ij} \nabla_{\bm{x}}J(T_{ij}(\bm{x}), y)\big|_{\bm{x}=\hat{\bm{x}}}\\
       =& \sum_{i,j} w_{ij} \big(\nabla_{T_{ij}(\bm{x})}J(T_{ij}(\bm{x}), y) \cdot \frac{\partial T_{ij}(\bm{x})}{\partial \bm{x}}\big) \Big|_{\bm{x}=\hat{\bm{x}}} \\
       =& \sum_{i,j} w_{ij} T_{-i-j}\big(\nabla_{\bm{x}}J(\bm{x}, y)\big|_{\bm{x}=T_{ij}(\hat{\bm{x}})}\big) \\
       \approx& \sum_{i,j} w_{ij}T_{-i-j}(\nabla_{\bm{x}}J(\bm{x}, y)\big|_{\bm{x}=\hat{\bm{x}}}).
\end{split}
\label{eq:gradient-1}
\end{equation}
Given Eq.~\eqref{eq:gradient-1}, we do not need to calculate the gradients for $(2k+1)^2$ images. Instead, we only need to get the gradient at the untranslated image $\hat{\bm{x}}$ and then average all the shifted gradients. This procedure is equivalent to convolving the gradient with a kernel composed of all the weights $w_{ij}$ as
\begin{equation*}
    \sum_{i,j} w_{ij}T_{-i-j}(\nabla_{\bm{x}}J(\bm{x}, y)\big|_{\bm{x}=\hat{\bm{x}}}) \Leftrightarrow \bm{W} * \nabla_{\bm{x}}J(\bm{x}, y)\big|_{\bm{x}=\hat{\bm{x}}},
    \label{eq:gradient-4}
\end{equation*}
where $\bm{W}$ is the kernel matrix of size $(2k+1)\times(2k+1)$, with $W_{i,j}=w_{-i-j}$.
We will specify $\bm{W}$ in the next section.

\vspace{-1.8ex}
\subsubsection{Kernel Matrix}
\label{sec:kernel}
There are many options to generate the kernel matrix $\bm{W}$.
A basic design principle is that the images with bigger shifts should have relatively lower weights to make the adversarial perturbation fool the model at the untranslated image effectively.
In this paper, we consider three different choices:
\begin{itemize}
\addtolength{\itemsep}{-1ex}
\item[(1)] A uniform kernel that $W_{i,j} = \nicefrac{1}{(2k+1)^2}$;
\item[(2)] A linear kernel that $\tilde{W}_{i,j} = (1-\nicefrac{|i|}{k+1})\cdot(1-\nicefrac{|j|}{k+1})$, and $W_{i,j}=\nicefrac{\tilde{W}_{i,j}}{\sum_{i,j}\tilde{W}_{i,j}}$;
\item[(3)] A Gaussian kernel that $\tilde{W}_{i,j} = \frac{1}{2\pi\sigma^2}\exp(-\frac{i^2+j^2}{2\sigma^2})$ where the standard deviation $\sigma=\nicefrac{k}{\sqrt{3}}$ to make the radius of the kernel be $3\sigma$, and $W_{i,j}=\nicefrac{\tilde{W}_{i,j}}{\sum_{i,j}\tilde{W}_{i,j}}$.
\end{itemize}
We will empirically compare the three kernels in Sec.~\ref{sec:kernel-exp}.

\vspace{-1.8ex}
\subsubsection{Attack Algorithms}
Note that in Sec.~\ref{sec:gradient}, we only illustrate how to calculate the gradient of the loss function defined in Eq.~\eqref{eq:problem-new}, but do not specify the update algorithm for generating adversarial examples. This indicates that our method can be integrated into any gradient-based attack method, \eg, FGSM, BIM, MI-FGSM, \etc.
For gradient-based attack methods presented in Sec.~\ref{sec:3.1}, in each step we calculate the gradient $\nabla_{\bm{x}}J(\bm{x}^{adv}_t, y)$  at the current solution $\bm{x}^{adv}_t$, then convolve the gradient with the pre-defined kernel $\bm{W}$, and finally obtain the new solution $\bm{x}^{adv}_{t+1}$ following the update rule in different attack methods.
For example, the combination of our translation-invariant method and the fast gradient sign method~\cite{goodfellow2014explaining} (TI-FGSM) has the following update rule
\begin{equation}
    \bm{x}^{adv} = \bm{x}^{real} + \epsilon\cdot\mathrm{sign}(\bm{W} *  \nabla_{\bm{x}}J(\bm{x}^{real},y)).
\end{equation}
Also, the integration of the translation-invariant method into the basic iterative method~\cite{kurakin2016adversarial} yields the TI-BIM algorithm
\begin{equation}
\bm{x}_{t+1}^{adv} = \bm{x}_t^{adv} + \alpha\cdot\mathrm{sign}(\bm{W} * \nabla_{\bm{x}}J(\bm{x}_t^{adv},y)).
\label{eq:iter}
\end{equation}
The translation-invariant method can be similarly integrated into MI-FGSM~\cite{Dong_2018_CVPR} and DIM~\cite{Xie2018Improving} as TI-MI-FGSM and TI-DIM, respectively.

\section{Experiments}
\label{sec:exp}

In this section, we present the experimental results to demonstrate the effectiveness of the proposed method.
We first specify the experimental settings in Sec.~\ref{sec:setup}.
Then we validate the translation-invariant property of convolutional neural networks in Sec.~\ref{sec:assumption}.
We further conduct two experiments to study the effects of different kernels and size of kernels in Sec.~\ref{sec:kernel-exp} and Sec.~\ref{sec:kernel-size}.
We finally compare the results of the proposed method with baseline methods in Sec.~\ref{sec:single} and Sec.~\ref{sec:ensemble}.

\subsection{Experimental Settings}
\label{sec:setup}

We use an ImageNet-compatible dataset\footnote{\url{https://github.com/tensorflow/cleverhans/tree/master/examples/nips17_adversarial_competition/dataset}} comprised of 1,000 images to conduct experiments. This dataset was used in the NIPS 2017 adversarial competition.
We include eight defense models which are shown to be robust against black-box attacks on the ImageNet dataset. These are 
\begin{itemize}
\vspace{-1.2ex}
\item Inc-v3\textsubscript{ens3}, Inc-v3\textsubscript{ens4}, IncRes-v2\textsubscript{ens}~\cite{tramer2017ensemble};
\vspace{-1.2ex}
\item high-level representation guided denoiser (HGD, rank-1 submission in the NIPS 2017 defense competition)~\cite{Liao2017Defense};
\vspace{-1.2ex}
\item input transformation through random resizing and padding (R\&P, rank-2 submission in the NIPS 2017 defense competition)~\cite{Xie2018Mitigating};
\vspace{-1.2ex}
\item input transformation through JPEG compression or total variance minimization (TVM) \cite{Guo2017Countering};
\vspace{-1.2ex}
\item rank-3 submission\footnote{\url{https://github.com/anlthms/nips-2017/tree/master/mmd}} in the NIPS 2017 defense competition (NIPS-r3).
\vspace{-1.2ex}
\end{itemize}
To attack these defenses based on the transferability, we also include four normally trained models---Inception v3 (Inc-v3)~\cite{szegedy2015rethinking}, Inception v4 (Inc-v4), Inception ResNet v2 (IncRes-v2)~\cite{szegedy2017inception}, and ResNet v2-152 (Res-v2-152)~\cite{he2016identity}, as the white-box models to generate adversarial examples.

\begin{figure}[t]
\centering
\includegraphics[width=1.0\columnwidth]{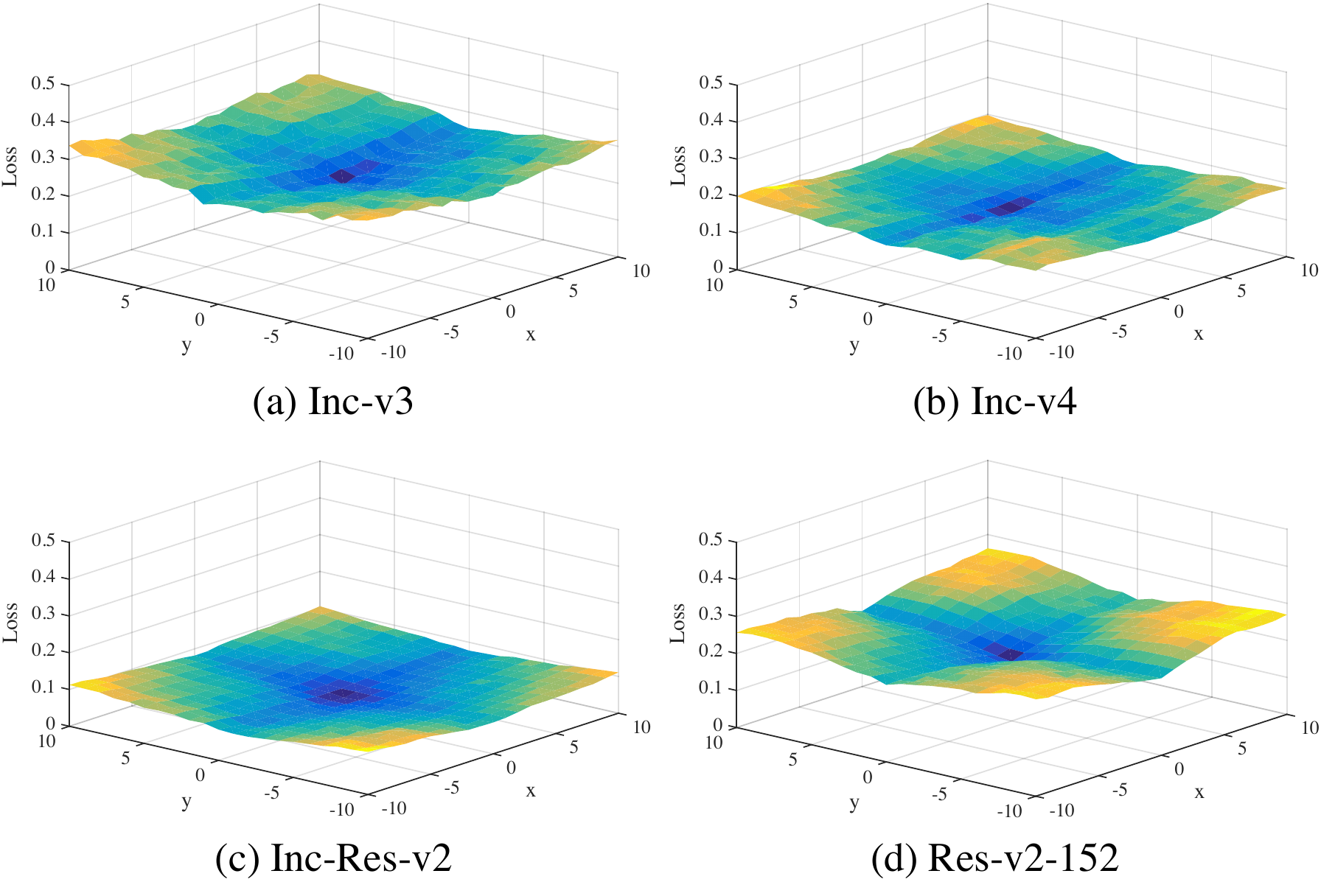}
\caption{We show the loss surfaces of Inc-v3, Inc-v4, IncRes-v2, and Res-v2-152 given the translated images at each position.}
\label{fig:loss}
\vspace{-2ex}
\end{figure}

\begin{table*}[!t]
\begin{center}
\footnotesize
\begin{tabular}{c|c|p{10ex}<{\centering}|p{10ex}<{\centering}|p{12ex}<{\centering}|p{8ex}<{\centering}|p{8ex}<{\centering}|p{8ex}<{\centering}|p{8ex}<{\centering}|p{8ex}<{\centering}}
\hline
& Attack & Inc-v3\textsubscript{ens3} & Inc-v3\textsubscript{ens4} & IncRes-v2\textsubscript{ens} & HGD & R\&P & JPEG & TVM & NIPS-r3 \\
\hline\hline
\multirow{3}{*}{TI-FGSM} & Uniform & 25.0 & 27.9 & 21.1 & 15.7 & 19.1 & 24.8 & 32.3 & 21.9 \\
                        & Linear & \bf30.7 & \bf32.4 & \bf24.2 & \bf20.9 & \bf23.3 & \bf28.1 & \bf34.6 & \bf25.8 \\
                        & Gaussian & 28.2 & 28.9 & 22.3 & 18.4 & 19.8 & 25.5 & 30.7 & 24.5 \\
\hline
\multirow{3}{*}{TI-MI-FGSM} & Uniform & 30.0 & 32.2 & 22.8 & 21.7 & 22.8 & 26.4 & 32.7 & 25.9 \\
                        & Linear & \bf35.8 & 35.0 & \bf26.8 & 25.5 & 23.4 & \bf29.0 & \bf35.8  & \bf27.5 \\
                        & Gaussian & \bf35.8 & \bf35.1 & 25.8 & \bf25.7 & \bf23.9 & 28.2 & 34.9 & 26.7 \\
\hline
\multirow{3}{*}{TI-DIM} & Uniform & 32.6 & 34.6 & 25.6 & 24.1 & 27.2 & 30.2 & 34.9 & 28.8 \\
                        & Linear & 45.2 & 47.0 & 34.9 & 35.6 & 35.2 & \bf38.5 & 43.6 & 39.7 \\
                        & Gaussian & \bf46.9 & \bf47.1 & \bf37.4 & \bf38.3 & \bf36.8 & 37.0 & \bf44.2 & \bf41.4 \\
\hline
\end{tabular}
\end{center}
\vspace{-1ex}
\caption{The success rates (\%) of black-box attacks against eight defenses with different choices of kernels. The adversarial examples are crafted for Inc-v3 by TI-FGSM, TI-MI-FGSM and TI-DIM with the uniform kernel, the linear kernel, and the Gaussian kernel, respectively.}
\label{tab:kernels}
\vspace{-1ex}
\end{table*}

\begin{figure*}[t]
\centering
\includegraphics[width=0.9\linewidth]{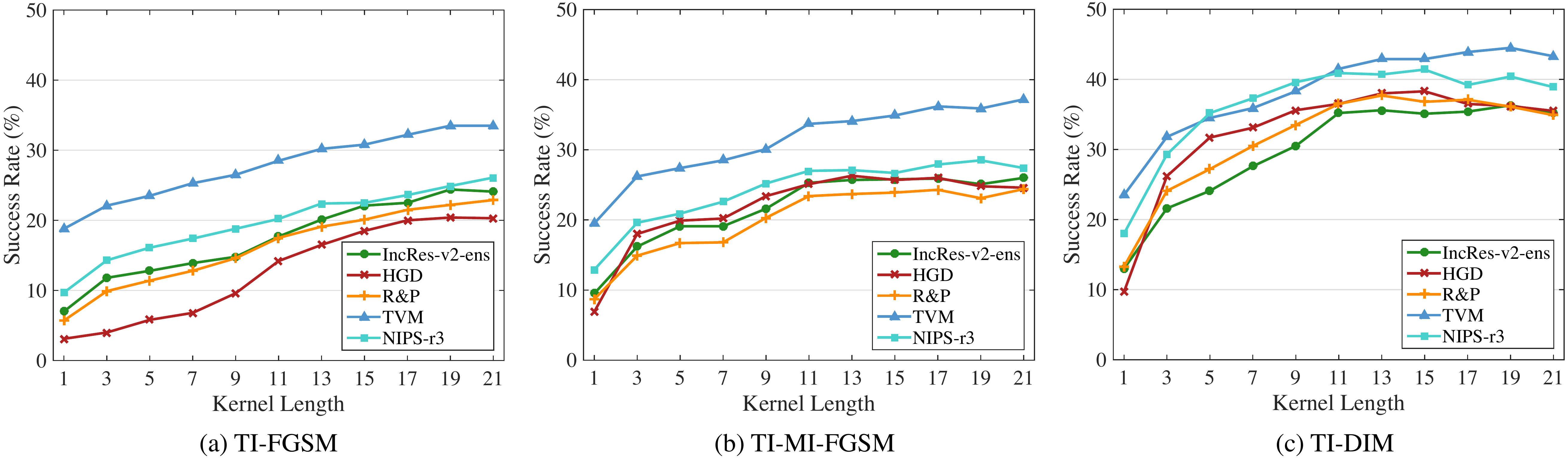}
\caption{The success rates (\%) of black-box attacks against IncRes-v2\textsubscript{ens}, HGD, R\&P, TVM, and NIPS-r3. The adversarial examples are generated for Inc-v3 with the kernel length ranging from $1$ to $21$.}
\label{fig:kernel}
\vspace{-1ex}
\end{figure*}

In our experiments, we integrate our method into the fast gradient sign method (FGSM)~\cite{goodfellow2014explaining}, momentum iterative fast gradient sign method (MI-FGSM)~\cite{Dong_2018_CVPR}, and diverse inputs method (DIM)~\cite{Xie2018Improving}. We do not include the basic iterative method~\cite{kurakin2016adversarial} and C\&W's method~\cite{carlini2017towards} since that they are not good at generating transferable adversarial examples~\cite{Dong_2018_CVPR}. We denote the attacks combined with our translation-invariant method as TI-FGSM, TI-MI-FGSM, and TI-DIM, respectively.

For the settings of hyper-parameters, we set the maximum perturbation to be $\epsilon=16$ among all experiments with pixel values in $[0,255]$. For the iterative attack methods, we set the number of iteration as $10$ and the step size as $\alpha=1.6$. 
For MI-FGSM and TI-MI-FGSM, we adopt the default decay factor $\mu=1.0$.
For DIM and TI-DIM, the transformation probability is set to $0.7$.
Please note that the settings for each attack method and its translation-invariant version are the same, because our method is not concerned with the specific attack procedure.

\subsection{Translation-Invariant Property of CNNs}
\label{sec:assumption}

We first verify the translation-invariant property of convolutional neural networks in this section.
We use the original 1,000 images from the dataset and shift them by $-10$ to $10$ pixels in each dimension. We input the original images as well as the translated images into Inc-v3, Inc-v4, IncRes-v2, and Res-v2-152, respectively.
The loss of each input image is given by the models. We average the loss over all translated images at each position, and show the loss surfaces in Fig.~\ref{fig:loss}.

It can be seen that the loss surfaces are generally smooth with the translations going from $-10$ to $10$ in each dimension.
So we could make the assumption that the translation-invariant property is almost held within a small range.
In our attacks, the images are shifted by no more than $10$ pixels along each dimension. The loss values would be very similar for the original and translated images. Therefore, we regard that a translated image is almost the same as the corresponding original image as inputs to the models.

\subsection{The Results of Different Kernels}
\label{sec:kernel-exp}

In the section, we show the experimental results of the proposed translation-invariant attack method with different choices of kernels.
We attack the Inc-v3 model by TI-FGSM, TI-MI-FGSM, and TI-DIM with three types of kernels, \ie, uniform kernel, linear kernel, and Gaussian kernel, as introduced in Sec.~\ref{sec:kernel}.
In Table~\ref{tab:kernels}, we report the success rates of black-box attacks against the eight defense models we study, where the success rates are the misclassification rates of the corresponding defense models with the generated adversarial images as inputs.

We can see that for TI-FGSM, the linear kernel leads to better results than the uniform kernel and the Gaussian kernel. And for more powerful attacks such as TI-MI-FGSM and TI-DIM, the Gaussian kernel achieves similar or even better results than the linear kernel. However, both of the linear kernel and the Gaussian kernel are more effective than the uniform kernel. It indicates that we should design the kernel that has lower weights for bigger shifts, as discussed in Sec.~\ref{sec:kernel}. We simply adopt the Gaussian kernel in the following experiments.

\begin{table*}[t]
\begin{center}
\footnotesize
\begin{tabular}{c|c|p{10ex}<{\centering}|p{10ex}<{\centering}|p{12ex}<{\centering}|p{8ex}<{\centering}|p{8ex}<{\centering}|p{8ex}<{\centering}|p{8ex}<{\centering}|p{8ex}<{\centering}}
\hline
& Attack & Inc-v3\textsubscript{ens3} & Inc-v3\textsubscript{ens4} & IncRes-v2\textsubscript{ens} & HGD & R\&P & JPEG & TVM & NIPS-r3 \\
\hline\hline
\multirow{2}{*}{Inc-v3} & FGSM & 15.6 & 14.7 & 7.0 & 2.1 & 6.5 & 19.9 & 18.8 & 9.8 \\
                        & TI-FGSM & \bf28.2 & \bf28.9 & \bf22.3 & \bf18.4 & \bf19.8 & \bf25.5 & \bf30.7 & \bf24.5 \\
\hline
\multirow{2}{*}{Inc-v4} & FGSM & 16.2 & 16.1 & 9.0 & 2.6 & 7.9 & 21.8 & 19.9 & 11.5 \\
                        & TI-FGSM & \bf28.2 & \bf28.3 & \bf21.4 & \bf18.1 & \bf21.6 & \bf27.9 & \bf31.8 & \bf24.6 \\
\hline
\multirow{2}{*}{IncRes-v2} & FGSM & 18.0 & 17.2 & 10.2 & 3.9 & 9.9 & 24.7 & 23.4 & 13.3 \\
                        & TI-FGSM & \bf32.8 & \bf33.6 & \bf28.1 & \bf25.4 & \bf28.1 & \bf32.4 & \bf38.5 & \bf31.4 \\
\hline
\multirow{2}{*}{Res-v2-152} & FGSM & 20.2 & 17.7 & 9.9 & 3.6 & 8.6 & 24.0 & 22.0 & 12.5 \\
                        & TI-FGSM & \bf34.6 & \bf34.5 & \bf27.8 & \bf24.4 & \bf27.4 & \bf32.7 & \bf38.1 & \bf30.1 \\
\hline
\end{tabular}
\end{center}
\vspace{-1ex}
\caption{The success rates (\%) of black-box attacks against eight defenses. The adversarial examples are crafted for Inc-v3, Inc-v4, IncRes-v2, and Res-v2-152 respectively using FGSM and TI-FGSM.}
\label{tab:fgsm}
\vspace{-1ex}
\end{table*}

\begin{table*}[t]
\begin{center}
\footnotesize
\begin{tabular}{c|c|p{10ex}<{\centering}|p{10ex}<{\centering}|p{12ex}<{\centering}|p{8ex}<{\centering}|p{8ex}<{\centering}|p{8ex}<{\centering}|p{8ex}<{\centering}|p{8ex}<{\centering}}
\hline
& Attack & Inc-v3\textsubscript{ens3} & Inc-v3\textsubscript{ens4} & IncRes-v2\textsubscript{ens} & HGD & R\&P & JPEG & TVM & NIPS-r3 \\
\hline\hline
\multirow{2}{*}{Inc-v3} & MI-FGSM & 20.5 & 17.4 & 9.5 & 6.9 & 8.7 & 20.3 & 19.4 & 12.9 \\
                        & TI-MI-FGSM & \bf35.8 & \bf35.1 & \bf25.8 & \bf25.7 & \bf23.9 & \bf28.2 & \bf34.9 & \bf26.7 \\
\hline
\multirow{2}{*}{Inc-v4} & MI-FGSM & 22.1 & 20.1 & 12.1 & 9.6 & 12.1 & 26.0 & 24.8 & 15.6 \\
                        & TI-MI-FGSM & \bf36.7 & \bf39.2 & \bf28.7 & \bf27.8 & \bf28.0 & \bf31.6 & \bf38.4 & \bf29.5 \\
\hline
\multirow{2}{*}{IncRes-v2} & MI-FGSM & 31.3 & 27.2 & 19.7 & 19.6 & 18.6 & 31.6 & 34.4 & 22.7 \\
                        & TI-MI-FGSM & \bf50.7 & \bf51.7 & \bf49.3 & \bf45.1 & \bf45.2 & \bf 45.9 & \bf55.4 & \bf46.2 \\
\hline
\multirow{2}{*}{Res-v2-152} & MI-FGSM & 25.1 & 23.7 & 13.3 & 15.1 & 14.6 & 31.2 & 24.5 & 18.0 \\
                        & TI-MI-FGSM & \bf39.9 & \bf37.7 & \bf32.8 & \bf31.8 & \bf31.1 & \bf 38.3 & \bf41.2 & \bf34.4 \\
\hline
\end{tabular}
\end{center}
\vspace{-1ex}
\caption{The success rates (\%) of black-box attacks against eight defenses. The adversarial examples are crafted for Inc-v3, Inc-v4, IncRes-v2, and Res-v2-152 respectively using MI-FGSM and TI-MI-FGSM.}

\label{tab:mim}
\vspace{-1ex}
\end{table*}

\begin{figure*}[!t]
\centering
\includegraphics[width=1.0\linewidth]{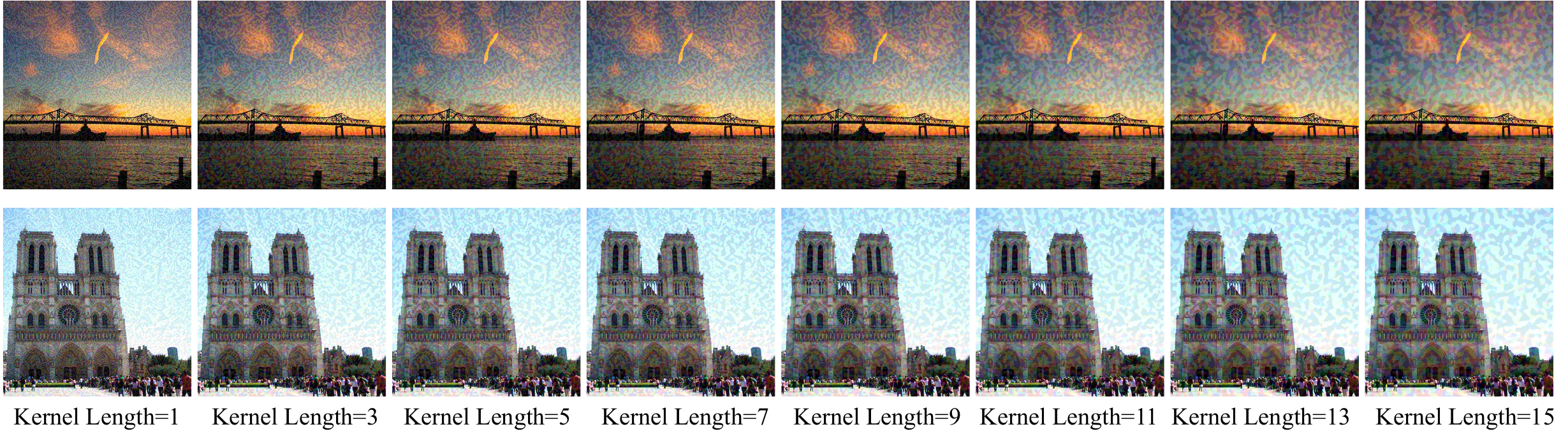}
\caption{The adversarial examples generated for Inc-v3 by TI-FGSM with different kernel sizes.}
\label{fig:kernel-image}
\vspace{-1ex}
\end{figure*}

\subsection{The Effect of Kernel Size}
\label{sec:kernel-size}

The size of the kernel $\bm{W}$ also plays a key role for improving the success rates of black-box attacks. If the kernel size equals to $1 \times 1$, the translation-invariant based attacks degenerate to their vanilla versions. Therefore, we conduct an ablation study to examine the effect of kernel sizes.

We attack the Inc-v3 model by TI-FGSM, TI-MI-FGSM, and TI-DIM with the Gaussian kernel, whose length ranges from $1$ to $21$ with a granularity $2$. In Fig.~\ref{fig:kernel}, we show the success rates against five defense models---IncRes-v2\textsubscript{ens}, HGD, R\&P, TVM, and NIPS-r3. The success rate continues increasing at first, and turns to remain stable after the kernel size exceeds $15\times 15$.
Therefore, the size of the kernel is set to $15\times 15$ in the following.

We also show the adversarial images generated for the Inc-v3 model by TI-FGSM with different kernel sizes in Fig.~\ref{fig:kernel-image}. Due to the smooth effect given by the kernel, we can see that the adversarial perturbations are smoother when using a bigger kernel.

\subsection{Single-Model Attacks}

\label{sec:single}

\begin{table*}[t]
\begin{center}
\footnotesize
\begin{tabular}{c|c|p{10ex}<{\centering}|p{10ex}<{\centering}|p{12ex}<{\centering}|p{8ex}<{\centering}|p{8ex}<{\centering}|p{8ex}<{\centering}|p{8ex}<{\centering}|p{8ex}<{\centering}}
\hline
& Attack & Inc-v3\textsubscript{ens3} & Inc-v3\textsubscript{ens4} & IncRes-v2\textsubscript{ens} & HGD & R\&P & JPEG & TVM & NIPS-r3 \\
\hline\hline
\multirow{2}{*}{Inc-v3} & DIM & 24.2 & 24.3 & 13.0 & 9.7 & 13.3 & 30.7 & 24.4 & 18.0 \\
                        & TI-DIM & \bf46.9 & \bf47.1 & \bf37.4 & \bf38.3 & \bf36.8 & \bf37.0 & \bf44.2 & \bf41.4 \\
\hline
\multirow{2}{*}{Inc-v4} & DIM & 28.3 & 27.5 & 15.6 & 14.6 & 17.2 & 38.6 & 29.1 & 14.1 \\
                        & TI-DIM & \bf48.6 & \bf47.5 & \bf38.7 & \bf40.3 & \bf39.3 & \bf43.5 & \bf45.6 & \bf41.9 \\
\hline
\multirow{2}{*}{IncRes-v2} & DIM & 41.2 & 40.0 & 27.9 & 32.4 & 30.2 & 47.2 & 41.7 & 37.6 \\
                        & TI-DIM & \bf61.3 & \bf60.1 & \bf59.5 & \bf58.7 & \bf61.4 & \bf55.7 & \bf66.2 & \bf61.5 \\
\hline
\multirow{2}{*}{Res-v2-152} & DIM & 40.5 & 36.0 & 24.1 & 32.6 & 26.4 & 42.4 & 36.8 & 34.4 \\
                        & TI-DIM & \bf56.1 & \bf55.5 & \bf49.5 & \bf51.8 & \bf50.4 & \bf50.8 & \bf 55.7 & \bf52.9 \\
\hline
\end{tabular}
\end{center}
\vspace{-1ex}
\caption{The success rates (\%) of black-box attacks against eight defenses. The adversarial examples are crafted for Inc-v3, Inc-v4, IncRes-v2, and Res-v2-152 respectively using DIM and TI-DIM.}
\label{tab:dim}
\vspace{-1ex}
\end{table*}

\begin{table*}[!t]
\begin{center}
\footnotesize
\begin{tabular}{c|p{10ex}<{\centering}|p{10ex}<{\centering}|p{12ex}<{\centering}|p{8ex}<{\centering}|p{8ex}<{\centering}|p{8ex}<{\centering}|p{8ex}<{\centering}|p{8ex}<{\centering}}
\hline
Attack & Inc-v3\textsubscript{ens3} & Inc-v3\textsubscript{ens4} & IncRes-v2\textsubscript{ens} & HGD & R\&P & JPEG & TVM & NIPS-r3 \\
\hline\hline
FGSM & 27.5 & 23.7 & 13.4 & 4.9 & 13.8 & 38.1 & 30.0 & 19.8 \\
TI-FGSM & \bf39.1 & \bf38.8 & \bf31.6 & \bf29.9 & \bf31.2 & \bf43.3 & \bf39.8 & \bf33.9 \\
\hline
MI-FGSM & 50.5 & 48.3 & 32.8 & 38.6 & 32.8 & 67.7 & 50.1 & 43.9 \\
TI-MI-FGSM & \bf76.4 & \bf74.4 & \bf69.6 & \bf73.3 & \bf68.3 & \bf77.2 & \bf72.1 & \bf71.4 \\
\hline
DIM & 66.0 & 63.3 & 45.9 & 57.7 & 51.7 & 82.5 & 64.1 & 63.7 \\
TI-DIM & \bf84.8 & \bf82.7 & \bf78.0 & \bf82.6 & \bf81.4 & \bf83.4 & \bf79.8 & \bf83.1 \\
\hline
\end{tabular}
\end{center}
\vspace{-1ex}
\caption{The success rates (\%) of black-box attacks against eight defenses. The adversarial examples are crafted for the ensemble of Inc-v3, Inc-v4, IncRes-v2, and Res-v2-152 using FGSM, TI-FGSM, MI-FGSM, TI-MI-FGSM, DIM, and TI-DIM.}
\label{tab:ens}
\vspace{-2ex}
\end{table*}

In this section, we compare the black-box success rates of the translation-invariant based attacks with baseline attacks.
We first perform adversarial attacks for Inc-v3, Inc-v4, IncRes-v2, and Res-v2-152 respectively using FGSM, MI-FGSM, DIM, and their extensions by combining with the translation-invariant attack method as TI-FGSM, TI-MI-FGSM, and TI-DIM.
We adopt the $15\times 15$ Gaussian kernel in this set of experiments.
We then use the generated adversarial examples to attack the eight defense models we consider based only on the transferability. We report the success rates of black-box attacks in Table~\ref{tab:fgsm} for FGSM and TI-FGSM, Table~\ref{tab:mim} for MI-FGSM and TI-MI-FGSM, and Table~\ref{tab:dim} for DIM and TI-DIM.

From the tables, we observe that the success rates against the defenses are improved by a large margin when using the proposed method regardless of the attack algorithms or the white-box models being attacked. 
In general, the translation-invariant based attacks consistently outperform the baseline attacks by $5\%\sim30\%$.
In particular, when using TI-DIM, the combination of our method and DIM, to attack the IncRes-v2 model, the resultant adversarial examples have about $60\%$ success rates against the defenses (as shown in Table~\ref{tab:dim}).
It demonstrates the vulnerability of the current defenses against black-box attacks.
The results also validate the effectiveness of the proposed method.
Although we only compare the results of our attack method with baseline methods against the defense models, our attacks remain the success rates of baseline attacks in the white-box setting and the black-box setting against normally trained models, which will be shown in the Appendix.
 
We show two adversarial images generated for the Inc-v3 model by FGSM and TI-FGSM in Fig.~\ref{fig:adv-images}. 
It can be seen that by using TI-FGSM, in which the gradients are convolved by a kernel $\bm{W}$ before applying to the raw images, the adversarial perturbations are much smoother than those generated by FGSM. The smooth effect also exists in other translation-invariant based attacks.

\subsection{Ensemble-based Attacks}
\label{sec:ensemble}

In this section, we further present the results when adversarial examples are generated for an ensemble of models. Liu \etal~\cite{liu2016delving} have shown that attacking multiple models at the same time can improve the transferability of the generated adversarial examples. It is due to that if an example remains adversarial for multiple models, it is more likely to transfer to another black-box model.

We adopt the ensemble method proposed in~\cite{Dong_2018_CVPR}, which fuses the logit activations of different models. We attack the ensemble of Inc-v3, Inc-v4, IncRes-v2, and Res-v2-152 with equal ensemble weights using FGSM, TI-FGSM, MI-FGSM, TI-MI-FGSM, DIM, and TI-DIM respectively. We also use the $15 \times 15$ Gaussian kernel in the translation-invariant based attacks.

In Table~\ref{tab:ens}, we show the results of black-box attacks against the eight defenses.
The proposed method also improves the success rates across all experiments over the baseline attacks.
It should be noted that \textit{the adversarial examples generated by TI-DIM can fool the state-of-the-art defenses at an $82\%$ success rate on average based on the transferability}. And the adversarial examples are generated for normally trained models unaware of the defense strategies.
The results in the paper demonstrate that the current defenses are far from real security, and cannot be deployed in real-world applications.

\vspace{-1ex}
\section{Conclusion}
\label{sec:conclusion}
\vspace{-1ex}

In this paper, we proposed a translation-invariant attack method to generate adversarial examples that are less sensitive to the discriminative regions of the white-box model being attacked, and have higher transferability against the defense models.
Our method optimizes an adversarial image by using a set of translated images. Based on an assumption, our method is efficiently implemented by convolving the gradient with a pre-defined kernel, and can be integrated into any gradient-based attack method. We conducted experiments to validate the effectiveness of the proposed method. Our best attack, TI-DIM, the combination of the proposed translation-invariant method and diverse inputs method~\cite{Xie2018Improving}, can fool eight state-of-the-art defenses at an $82\%$ success rate on average, where the adversarial examples are generated against four normally trained models. The results identify the vulnerability of the current defenses, and thus raise security issues for the development of more robust deep learning models.
We make our codes public at \url{https://github.com/dongyp13/Translation-Invariant-Attacks}.

\vspace{-1ex}
\section*{Acknowledgements}
\vspace{-1ex}
This work was supported by the National Key Research and Development Program of China (No. 2017YFA0700904), NSFC Projects (Nos. 61620106010, 61621136008, 61571261), Beijing NSF Project (No. L172037), DITD Program JCKY2017204B064, Tiangong Institute for Intelligent Computing, NVIDIA NVAIL Program, and the projects from Siemens and Intel.

{\small
\bibliographystyle{ieee}
\bibliography{egbib}
}

\clearpage
\setlength{\tabcolsep}{4pt}
\noindent \begin{center} {\large  \textbf{Appendix}} \end{center}

We first show the results of the proposed translation-invariant attack method for white-box attacks and black-box attacks against normally trained models.
We adopt the same settings for attacks. We also generate adversarial examples for Inception v3 (Inc-v3)~\cite{szegedy2015rethinking}, Inception v4 (Inc-v4), Inception ResNet v2 (IncRes-v2)~\cite{szegedy2017inception}, and ResNet v2-152 (Res-v2-152)~\cite{he2016identity}, respectively, using FGSM, TI-FGSM, MI-FGSM, TI-MI-FGSM, DIM, and TI-DIM.
For the translation-invariant based attacks, we use the $7\times7$ Gaussian kernel, since that the normally trained models have similar discriminative regions.
We then use these adversarial examples to attack six normally trained models---Inc-v3, Inc-v4, IncRes-v2, Res-v2-152, VGG-16~\cite{simonyan2014very}, and Res-v1-152~\cite{he2015deep}. The results are shown in Table~\ref{tab:fgsm-2} for FGSM and TI-FGSM, Table~\ref{tab:mim-2} for MI-FGSM and TI-MI-FGSM, and Table~\ref{tab:dim-2} for DIM and TI-DIM. The translation-invariant based attacks get better results in most cases than the baseline attacks.

Moreover, the experiments above and in the main paper are conducted based on the $L_{\infty}$ norm bound. We further demonstrate the applicability of the proposed method for other norm bounds, especially the $L_2$ norm bound.
Similar to the results in Table 2-5, we present the results of FGSM and TI-FGSM in Table~\ref{tab:fgsm-3}, MI-FGSM and TI-MI-FGSM in Table~\ref{tab:mim-3}, DIM and TI-DIM in Table~\ref{tab:dim-3}, and the ensemble method in Table~\ref{tab:ens-3}.
All those results are based on the $L_2$ norm bound, and we set the maximum perturbation $\epsilon=10\cdot\sqrt{d}$, where $d$ is the dimension of input images.
The results based on the $L_2$ norm bound also show the effectiveness of the proposed method.

\begin{table*}[b]
\begin{center}
\begin{tabular}{c|c|p{12ex}<{\centering}|p{12ex}<{\centering}|p{12ex}<{\centering}|p{12ex}<{\centering}|p{12ex}<{\centering}|p{12ex}<{\centering}}
\hline
& Attack & Inc-v3 & Inc-v4 & IncRes-v2 & Res-v2-152 & VGG-16 & Res-v1-152 \\
\hline\hline
\multirow{2}{*}{Inc-v3} & FGSM & \bf79.6* & 35.9 & 30.6 & 30.2 & 49.7 & 36.3  \\
                        & TI-FGSM & 75.4* & \bf37.3 & \bf32.1 & \bf34.1 & \bf62.0 & \bf44.9  \\
\hline
\multirow{2}{*}{Inc-v4} & FGSM & 43.1 & \bf72.6* & 32.5 & 34.3 & 50.7 & 37.7  \\
                        & TI-FGSM & \bf45.3 & 68.1* & \bf33.7 & \bf35.4 & \bf63.3 & \bf46.2 \\
\hline
\multirow{2}{*}{IncRes-v2} & FGSM & 44.3 & 36.1 & \bf64.3* & 31.9 & 49.4 & 38.6 \\
                        & TI-FGSM & \bf49.7 & \bf41.5 & 63.7* & \bf40.1 & \bf64.2 & \bf46.7 \\
\hline
\multirow{2}{*}{Res-v2-152} & FGSM & 40.1 & 34.0 & 30.3 & \bf81.3* & 50.5 & 40.8 \\
                        & TI-FGSM & \bf46.4 & \bf39.3 & \bf33.4 & 78.9* & \bf64.7 & \bf50.4\\
\hline
\end{tabular}
\end{center}
\caption{The success rates (\%) of adversarial attacks against six normally trained models---Inc-v3, Inc-v4, IncRes-v2, Res-v2-152, VGG-16, and Res-v1-152. The adversarial examples are crafted for Inc-v3, Inc-v4, IncRes-v2, and Res-v2-152, respectively, using FGSM and TI-FGSM. * indicates the white-box attacks.}
\label{tab:fgsm-2}
\end{table*}

\begin{table*}[b]
\begin{center}
\begin{tabular}{c|c|p{12ex}<{\centering}|p{12ex}<{\centering}|p{12ex}<{\centering}|p{12ex}<{\centering}|p{12ex}<{\centering}|p{12ex}<{\centering}}
\hline
& Attack & Inc-v3 & Inc-v4 & IncRes-v2 & Res-v2-152 & VGG-16 & Res-v1-152 \\
\hline\hline
\multirow{2}{*}{Inc-v3} & MI-FGSM & 97.8* & 47.1 & 46.4 & 38.7 & 50.3 & 38.1  \\
                        & TI-MI-FGSM & \bf97.9* & \bf52.4 & \bf47.9 & \bf41.1 & \bf63.4 & \bf48.1  \\
\hline
\multirow{2}{*}{Inc-v4} & MI-FGSM & 67.1 & \bf98.8* & 54.3 & 47.0 & 58.5 & 43.2  \\
                        & TI-MI-FGSM & \bf68.6 & \bf98.8* & \bf55.3 & \bf47.7 & \bf69.0 & \bf51.3 \\
\hline
\multirow{2}{*}{IncRes-v2} & MI-FGSM & 74.8 & 64.8 & \bf100.0* & 54.5 & 59.3 & 50.8 \\
                        & TI-MI-FGSM & \bf76.1 & \bf69.5 & \bf100.0* & \bf59.6 & \bf74.4 & \bf61.5 \\
\hline
\multirow{2}{*}{Res-v2-152} & MI-FGSM & 54.2 & 48.1 & 44.3 & \bf97.5* & 52.6 & 48.7 \\
                        & TI-MI-FGSM & \bf55.6 & \bf50.9 & \bf45.1 & 97.4* & \bf65.6 & \bf59.6\\
\hline
\end{tabular}
\end{center}
\caption{The success rates (\%) of adversarial attacks against six normally trained models---Inc-v3, Inc-v4, IncRes-v2, Res-v2-152, VGG-16, and Res-v1-152. The adversarial examples are crafted for Inc-v3, Inc-v4, IncRes-v2, and Res-v2-152, respectively, using MI-FGSM and TI-MI-FGSM. * indicates the white-box attacks.}
\label{tab:mim-2}
\end{table*}

\begin{table*}[!h]
\begin{center}
\begin{tabular}{c|c|p{12ex}<{\centering}|p{12ex}<{\centering}|p{12ex}<{\centering}|p{12ex}<{\centering}|p{12ex}<{\centering}|p{12ex}<{\centering}}
\hline
& Attack & Inc-v3 & Inc-v4 & IncRes-v2 & Res-v2-152 & VGG-16 & Res-v1-152 \\
\hline\hline
\multirow{2}{*}{Inc-v3} & DIM & 98.3* & 73.8 & 67.8 & 58.4 & 62.5 & 49.3  \\
                        & TI-DIM & \bf98.5* & \bf75.2 & \bf69.2 & \bf59.0 & \bf74.3 & \bf59.1  \\
\hline
\multirow{2}{*}{Inc-v4} & DIM & \bf81.8 & 98.2* & \bf74.2 & \bf65.1 & 65.5 & 51.4  \\
                        & TI-DIM & 80.7 & \bf98.7* & 73.2 & 62.7 & \bf77.4 & \bf59.8 \\
\hline
\multirow{2}{*}{IncRes-v2} & DIM & 86.1 & 83.5 & \bf99.1* & 73.5 & 67.9 & 62.7 \\
                        & TI-DIM & \bf86.4 & \bf85.5 & 98.8* & \bf76.3 & \bf79.3 & \bf72.2 \\
\hline
\multirow{2}{*}{Res-v2-152} & DIM & \bf77.0 & \bf77.8 & \bf73.5 & \bf97.4* & 67.4 & 67.8 \\
                        & TI-DIM & \bf77.0 & 73.9 & 73.2 & 97.2* & \bf78.4 & \bf77.8 \\
\hline
\end{tabular}
\end{center}
\caption{The success rates (\%) of adversarial attacks against six normally trained models---Inc-v3, Inc-v4, IncRes-v2, Res-v2-152, VGG-16, and Res-v1-152. The adversarial examples are crafted for Inc-v3, Inc-v4, IncRes-v2, and Res-v2-152, respectively, using DIM and TI-DIM. * indicates the white-box attacks.}
\label{tab:dim-2}
\end{table*}

\begin{table*}[t]
\begin{center}
\begin{tabular}{c|c|p{10ex}<{\centering}|p{10ex}<{\centering}|p{12ex}<{\centering}|p{8ex}<{\centering}|p{8ex}<{\centering}|p{8ex}<{\centering}|p{8ex}<{\centering}|p{8ex}<{\centering}}
\hline
& Attack & Inc-v3\textsubscript{ens3} & Inc-v3\textsubscript{ens4} & IncRes-v2\textsubscript{ens} & HGD & R\&P & JPEG & TVM & NIPS-r3 \\
\hline\hline
\multirow{2}{*}{Inc-v3} & FGSM & 13.7 & 14.5 & 6.8 & 6.0 & 6.1 & 10.9 & 22.0 & 8.2 \\
                        & TI-FGSM & \bf15.2 & \bf15.7 & \bf10.2 & \bf8.2 & \bf18.8 & \bf11.0 & \bf25.7 & \bf10.4 \\
\hline
\multirow{2}{*}{Inc-v4} & FGSM & \bf13.9 & 15.0 & 8.2 & \bf8.3 & 7.4 & \bf11.5 & 22.2 & 8.5 \\
                        & TI-FGSM & \bf13.9 & \bf16.2 & \bf10.4 & 8.0 & \bf9.1 & 11.3 & \bf24.3 & \bf8.9 \\
\hline
\multirow{2}{*}{IncRes-v2} & FGSM & 16.0 & 17.5 & 11.3 & 10.8 & 10.2 & 14.4 & 26.2 & 11.6 \\
                        & TI-FGSM & \bf18.1 & \bf18.5 & \bf15.5 & \bf12.3 & \bf13.2 & \bf14.7 & \bf29.4 & \bf13.6 \\
\hline
\multirow{2}{*}{Res-v2-152} & FGSM & 12.7 & 15.1 & 8.1 & 7.0 & 7.1 & 10.2 & 20.3 & 8.2 \\
                        & TI-FGSM & \bf13.4 & \bf15.8 & \bf9.7 & \bf7.2 & \bf7.9 & \bf10.7 & \bf22.5 & \bf9.1 \\
\hline
\end{tabular}
\end{center}
\vspace{-1ex}
\caption{The success rates (\%) of black-box attacks against eight defenses based on the $L_2$ norm bound. The adversarial examples are crafted for Inc-v3, Inc-v4, IncRes-v2, and Res-v2-152 respectively using FGSM and TI-FGSM.}
\label{tab:fgsm-3}
\end{table*}

\begin{table*}[t]
\begin{center}
\begin{tabular}{c|c|p{10ex}<{\centering}|p{10ex}<{\centering}|p{12ex}<{\centering}|p{8ex}<{\centering}|p{8ex}<{\centering}|p{8ex}<{\centering}|p{8ex}<{\centering}|p{8ex}<{\centering}}
\hline
& Attack & Inc-v3\textsubscript{ens3} & Inc-v3\textsubscript{ens4} & IncRes-v2\textsubscript{ens} & HGD & R\&P & JPEG & TVM & NIPS-r3 \\
\hline\hline
\multirow{2}{*}{Inc-v3} & MI-FGSM & 15.9 & 16.3 & 7.0 & 7.8 & 7.5 & 12.8 & 15.7 & 8.4 \\
                        & TI-MI-FGSM & \bf22.8 & \bf24.6 & \bf14.8 & \bf14.0 & \bf13.0 & \bf15.8 & \bf22.7 & \bf15.1 \\
\hline
\multirow{2}{*}{Inc-v4} & MI-FGSM & 18.1 & 18.7 & 8.3 & 9.3 & 9.0 & 14.9 & 17.5 & 10.7 \\
                        & TI-MI-FGSM & \bf24.3 & \bf25.5 & \bf27.9 & \bf15.7 & \bf15.9 & \bf29.0 & \bf25.2 & \bf16.5 \\
\hline
\multirow{2}{*}{IncRes-v2} & MI-FGSM & 22.9 & 21.6 & 16.6 & 17.1 & 15.2 & 22.2 & 20.9 & 18.0 \\
                        & TI-MI-FGSM & \bf35.0 & \bf35.8 & \bf30.5 & \bf26.3 & \bf26.4 & \bf29.8 & \bf35.6 & \bf28.8 \\
\hline
\multirow{2}{*}{Res-v2-152} & MI-FGSM & 18.6 & 18.7 & 10.4 & 12.4 & 10.8 & 14.9 & 15.9 & 11.1 \\
                        & TI-MI-FGSM & \bf21.6 & \bf23.3 & \bf17.3 & \bf15.1 & \bf15.6 & \bf18.7 & \bf24.6 & \bf17.6 \\
\hline
\end{tabular}
\end{center}
\caption{The success rates (\%) of black-box attacks against eight defenses based on the $L_2$ norm bound. The adversarial examples are crafted for Inc-v3, Inc-v4, IncRes-v2, and Res-v2-152 respectively using MI-FGSM and TI-MI-FGSM.}

\label{tab:mim-3}
\end{table*}

\begin{table*}[t]
\begin{center}
\begin{tabular}{c|c|p{10ex}<{\centering}|p{10ex}<{\centering}|p{12ex}<{\centering}|p{8ex}<{\centering}|p{8ex}<{\centering}|p{8ex}<{\centering}|p{8ex}<{\centering}|p{8ex}<{\centering}}
\hline
& Attack & Inc-v3\textsubscript{ens3} & Inc-v3\textsubscript{ens4} & IncRes-v2\textsubscript{ens} & HGD & R\&P & JPEG & TVM & NIPS-r3 \\
\hline\hline
\multirow{2}{*}{Inc-v3} & DIM & 17.9 & 21.8 & 9.7 & 11.8 & 10.0 & 15.5 & 17.0 & 12.7 \\
                        & TI-DIM & \bf29.6 & \bf31.9 & \bf22.0 & \bf20.1 & \bf20.0 & \bf22.0 & \bf27.3 & \bf23.9 \\
\hline
\multirow{2}{*}{Inc-v4} & DIM & 21.6 & 22.2 & 12.9 & 15.8 & 13.3 & 20.5 & 19.2 & 16.6 \\
                        & TI-DIM & \bf31.0 & \bf33.1 & \bf24.0 & \bf22.8 & \bf22.9 & \bf24.8 & \bf29.2 & \bf25.1 \\
\hline
\multirow{2}{*}{IncRes-v2} & DIM & 34.5 & 31.0 & 23.8 & 27.0 & 25.8 & 31.5 & 25.0 & 26.9 \\
                        & TI-DIM & \bf43.3 & \bf45.2 & \bf42.4 & \bf39.3 & \bf42.7 & \bf42.2 & \bf43.3 & \bf41.2 \\
\hline
\multirow{2}{*}{Res-v2-152} & DIM & 29.0 & 30.1 & 18.7 & 27.8 & 19.8 & 26.7 & 21.3 & 23.1 \\
                        & TI-DIM & \bf36.3 & \bf37.2 & \bf28.9 & \bf28.0 & \bf30.0 & \bf28.4 & \bf36.1 & \bf32.7 \\
\hline
\end{tabular}
\end{center}
\vspace{-1ex}
\caption{The success rates (\%) of black-box attacks against eight defenses based on the $L_2$ norm bound. The adversarial examples are crafted for Inc-v3, Inc-v4, IncRes-v2, and Res-v2-152 respectively using DIM and TI-DIM.}
\label{tab:dim-3}
\end{table*}

\begin{table*}[!t]
\begin{center}
\begin{tabular}{c|p{10ex}<{\centering}|p{10ex}<{\centering}|p{12ex}<{\centering}|p{8ex}<{\centering}|p{8ex}<{\centering}|p{8ex}<{\centering}|p{8ex}<{\centering}|p{8ex}<{\centering}}
\hline
Attack & Inc-v3\textsubscript{ens3} & Inc-v3\textsubscript{ens4} & IncRes-v2\textsubscript{ens} & HGD & R\&P & JPEG & TVM & NIPS-r3 \\
\hline\hline
FGSM & \bf26.6 & \bf27.3 & 16.0 & \bf18.1 & 16.5 & \bf21.1 & 23.7 & 17.9 \\
TI-FGSM & 26.1 & 26.7 & \bf19.2 & 17.1 & \bf19.1 & 20.0 & \bf27.2 & \bf19.1 \\
\hline
MI-FGSM & 44.3 & 42.8 & 27.2 & 40.7 & 28.1 & 43.6 & 30.8 & 34.4 \\
TI-MI-FGSM & \bf59.3 & \bf59.0 & \bf53.0 & \bf54.6 & \bf50.0 & \bf53.3 & \bf51.3 & \bf51.1 \\
\hline
DIM & 57.0 & 54.7 & 37.4 & 58.9 & 43.4 & \bf60.3 & 37.3 & 50.3 \\
TI-DIM & \bf66.9 & \bf66.0 & \bf60.4 & \bf63.2 & \bf62.9 & 58.4 & \bf58.4 & \bf62.7 \\
\hline
\end{tabular}
\end{center}
\caption{The success rates (\%) of black-box attacks against eight defenses based on the $L_2$ norm bound. The adversarial examples are crafted for the ensemble of Inc-v3, Inc-v4, IncRes-v2, and Res-v2-152 using FGSM, TI-FGSM, MI-FGSM, TI-MI-FGSM, DIM, and TI-DIM.}
\label{tab:ens-3}
\end{table*}

\end{document}